\documentclass{article}

\usepackage{times}
\usepackage{graphicx} % more modern
\usepackage{subfigure} 

\usepackage{natbib}

\usepackage[boxed]{algorithm2e}
\usepackage{algorithmic}
\usepackage{amsfonts}
\usepackage{amsmath}
\usepackage{amsthm}

\usepackage{hyperref}

\usepackage[accepted]{icml2015}

\graphicspath{{images/}}

\newtheorem{theorem}{Theorem}[section]
\newtheorem{lemma}{Lemma}[section]
\newtheorem{definition}{Definition}[section]

\icmltitlerunning{Fast Online Clustering with Randomized Skeleton Sets}

\begin{document} 

\twocolumn[
\icmltitle{Fast Online Clustering with Randomized Skeleton Sets}

% It is OKAY to include author information, even for blind
% submissions: the style file will automatically remove it for you
% unless you've provided the [accepted] option to the icml2015
% package.
\icmlauthor{Krzysztof Choromanski}{kchoro@google.com}
\icmladdress{Google Research,
             76 Ninth Ave, New York, NY 10011 USA}
\icmlauthor{Sanjiv Kumar}{sanjivk@google.com}
\icmladdress{Google Research,
             76 Ninth Ave, New York, NY 10011 USA}
\icmlauthor{Xiaofeng Liu}{xiaofengliu@google.com}
\icmladdress{Google Research,
             76 Ninth Ave, New York, NY 10011 USA}             

% You may provide any keywords that you 
% find helpful for describing your paper; these are used to populate 
% the "keywords" metadata in the PDF but will not be shown in the document
\icmlkeywords{nonparametric clustering, skeleton set, online clustering}

\vskip 0.3in
]

\begin{abstract} 
We present a new fast online clustering algorithm that reliably recovers arbitrary-shaped data clusters in high throughout
data streams. Unlike the existing state-of-the-art online clustering methods based on k-means or k-medoid, it does not
make any restrictive generative assumptions. In addition, in contrast to existing nonparametric clustering techniques such
as DBScan or DenStream, it gives provable theoretical guarantees. To achieve fast clustering, we
propose to represent each cluster by a skeleton set which is updated continuously as new data is seen. A skeleton set  consists of weighted samples from the data where
weights encode local densities. The size of each skeleton set is adapted according to the cluster geometry. The proposed technique automatically detects the
number of clusters and is robust to outliers. The algorithm
works for the infinite data stream where more than one pass over the data is not feasible. We provide
theoretical guarantees on the quality of the clustering and also demonstrate its advantage over the existing
state-of-the-art on several datasets.
\end{abstract} 

\section{Introduction}
\label{sec:intro}

Online clustering in massive data streams is becoming important as data in a variety of fields including social media,
finance and web applications arrives as a high throughput stream. In social networks, detecting and tracking clusters
or communities is important to analyze evolutionary patterns. Similarly, online clustering can lead to spam and fraud
detection in web applications such as detecting unusual mass activities in email services or online reviews. There
exist several challenges in developing a good clustering algorithm in a high throughput online scenario. In real-world
applications, the number and shape of the clusters is typically unknown. The existing state-of-the-art online clustering
methods with provable theoretical guarantees are primarily based on k-means or k-median/medoid, which assume apriori
knowledge of the number of clusters and inherently make strong generative assumptions about the clusters.
%\cite{zhang96birch, guha, ailon, meyerson}. 
These assumptions force the retrieved clusters to be convex, leading to
poor clustering for many real-world tasks.  There exist several nonparametric techniques that do not make simplistic generative assumptions, but are mostly based on heuristics and lack theoretical guarantees. 
Moreover, in a true online scenario one needs to deal with continuos
streams precluding the use of multiple passes over data applicable for finite-size streams commonly assumed by many
techniques.
% \cite{ailon, meyerson}. 
Potential drift in data distribution over time is another practical
difficulty that needs to be handled effectively. Finally, the clustering procedure should be efficient both in space
and time to be able to handle massive data streams.
% as well as should have sound theoretical guarantees on the clustering performance. 

%including DBScan \cite{ester}, DenStream algorithm \cite{cao} and different variants of the LeaderFollower algorithms \cite{shah} 

%either designed to work in the very specific setting \cite{strnadova} or are
%just heuristic approaches.

In this paper we propose a novel Skeleton-based Online Clustering (SOC) algorithm to address the above challenges.
The basic idea of SOC is to represent each cluster via a compact skeleton set which faithfully
captures the geometry of the cluster. Each skeleton set maintains a small random sample from the corresponding cluster in an
online fashion, which is updated fast using the new data points. Each skeleton point is weighted according to local
density around it. The number
of skeleton points is automatically adapted to the structure of the cluster in such a way
that more complicated shapes are approximated by more skeleton points. 
%Therefore the number of skeleton points of the given cluster can change over time as the cluster evolves. 
The skeleton sets are updated by a random procedure which
provides robustness in the presence of outliers. The proposed algorithm automatically recovers the
correct number of clusters in the data with high probability as more and more data is seen. The update strategy of the
skeleton sets also allows the clustering method to automatically adapt to any drift in data distribution. In SOC, clusters can be merged as well as split over time. 
%the algorithms can work in two modes. First, clusters are formed via lazy merging and splitting is not used, and second, both merging and splitting can happen. Splitting is
%performed on an undirected sparse graph defined on all skeleton points of the given clusters. This procedure is more expensive (thus is performed much
%less frequently than merging) but it allows for even more effective clusters' reconstruction since merging can be now
%performed more aggressively (even if two different clusters are merged by mistake the newly constructed cluster will be
%split soon after this happens). The SOC algorithm is quite easy to implement and gives good results for
%arbitrary-shaped clusters as shown in the experiments. 
We also provide theoretical guarantees on the quality of
clusters obtained from the proposed method.

\subsection{Related work}

%Clustering is a well-established field with huge literature, but most of these techniques work in the offline mode when all the data is available in a single shot. 
%In this work we are interested in online clustering, where data arrives as a stream (possibly infinite) and the goal 
%is to maintain a good clustering for the data seen so far.
In comparison to the huge literature in offline clustering, work on online clustering has been somewhat limited. 
Most of the existing online clustering algorithms that have theoretical guarantees fall under model-based techniques such as k-mean, k-median or k-medoid \cite{guha, ailon, meyerson, bagirov}. 
They assume specific shape of the clusters such as spheres that trivially leads to their compact representation using just a  few parameters, e.g., center, radius and the number of points. 
%The work in \cite{ailon} proposed
%an algorithm to approximate the k-means objective in the streaming scenario where only one pass over the entire data is allowed. In \cite{meyerson}, an improved alogorithm is proposed to do k-means for massive datasets that do not fit in the memory and have to be accessed sequentially. This technique uses multiple passes over data. 
However, as discussed before, these model based algorithms fail to capture arbitrary clusters in the data and can perform poorly.

There exist several nonparametric clustering methods where no assumption is made about the cluster shapes. Popular among them are DBScan \cite{ester}, CluStream \cite{aggarwal}, and 
DenStream  \cite{cao}.  Recent surveys have described several variants of these algorithms  \cite{silva, amini}. 
The DenStream and CluStream methods create microclusters based on local densities, and combine them to form bigger clusters over time. 
However, these methods need to periodically perform offline partitioning of all the microclusters to form the clusters, which is not 
suitable for online clustering of massive data streams. Leader-Follower algorithm is another popular method and there exist several variants of it \cite{duda}\cite{shah}.  
These techniques typically encode every cluster by one center which is updated continuously as new points belonging to the cluster are detected. Such a cluster 
representation is not rich enough to encode more complex clusters. Overall, the main drawback of the above nonparametric online clustering algorithms is that 
they are mostly based on heuristics and lack any theoretical guarantees. They also require extensive hand tuning of the parameters. In \cite{shah}, the authors 
assume each cluster to be clique in order to provide theoretical guarantees, which is  very restrictive in real-world.

Another popular method used in the context of incremental clustering is doubling algorithm \cite{charikar}. Its standard version encodes every
cluster by just one point. Furthermore, even though it allows for merging clusters, it does not permit to split them. 
We implement a variant of the method, where instead of one center several centers are kept per cluster. 
As we will show in experimental section, this purely deterministic approach, even though with some theoretical guarantees, is too
sensitive to outliers. 

Our proposed SOC algorithm shares a few similarities with two existing techniques: CURE algorithm \cite{rastogi}, and  \textit{core-set} \cite{indyk}. In CURE, similar to SOC, each cluster is represented by a random sample
of data instead of just one center to handle arbitrary cluster shapes. CURE, however, is an offline hierarchical agglomerative clustering approach with running time quadratic in the size of data, which is too slow for online applications. In core-set based clustering, the aim is to encode a complicated cluster shape via a compact sample of points.
The existing state-of-the-art algorithms that use the idea of the core-set \cite{gonzalez, alon, varadarajan, indyk} are  computationally too intensive to be useful for online clustering in practice. The running time 
%even though only inversely-proportional to the approximation precision $\epsilon$
%for the best variants of the method \cite{indyk}, 
is exponential in the number of stored skeleton points. Furthermore, the variants that give provable theoretical guarantees are inherently offline methods
that often require several passes over the data to produce good-quality clustering. For instance, the algorithm presented in \cite{indyk} needs to be 
rerun $2^{O(k)}\log(n)$ times, where $k$ is the size of the core-set an $n$ is dataset size. 

Nonparameteric graph-based techniques such as spectral clustering can recover arbitrary shaped clusters but they are appropriate mainly for the offline setting \cite{ng}. Moreover, they also assume a priori knowledge of the number of clusters. Several relaxations such as iterative biclustering have been proposed to overcome the need to know the number of clusters apriori  but these methods cannot be extended to an online setting. 
%Spectral clustering techniques are applied in many different settings (see: \cite{bach}) however they do not achieve any
%strict theoretical guarantees regarding the quality of the output partitioning.
Recently, there has been some work on incremental spectral clustering 
which essentially iteratively modifies the Graph Laplacian \cite{ning, langone, jia, song}. 
In a true online setting, however, even building a good initial Graph Laplacian is infeasible either due to the lack of enough data or the computational bottlenecks.
%
%The clustering problem becomes easier if one can assume something about the structure of the underlying data.
%The conditions may be purely combinatorial such as graph sparsity from \cite{chen} or the planted partition model
%analyzed in \cite{condon} or more analytical ones such as those from \cite{vidal}, where sparse manifold clustering is  analyzed.
%The online data may be assumed to be generated according to some ergodic process \cite{ryabko}. It may also be the case that data distribution
%changes over time. This leads to the so-called \textit{evolutionary clustering model} \cite{song}.
%Other models describe building a hierarchy of clusters to obtain optimal data partitioning \cite{krishna}.
%Even though many formulations of the clustering task are NP-hard, for some of them with extra assumptions, guarantees on clustering quality can be derived \cite{bansal}.

\section{Clustering Framework}
\label{sec:clustering}

In this work, since we are interested in retrieving arbitrary-shaped clusters, it is important to first define what constitutes a good
\textit{cluster}. The traditional model-based techniques assume a global distance measure (e.g., $L_2$ distance in
k-means), which restricts one to convex-shaped clusters. Instead, the proposed algorithm works in a
\textit{nonparametric} setting where clusters are defined by an intuitive notion of paths with 'short-edges'. Two
points are more likely to belong to the same cluster if there exist many paths in the data neighborhood graph such that
any two consecutive points on each path are not "far" from each other. Clearly, the overall distance between two
points can be large but they can still be in the same cluster. Such a setting enables us to consider many complicated
cluster shapes. To emphasize, the idea of the neighborhood graph is to understand the cluster definition we
implicitly utilize in this work. We do not need to explicitly construct a graph in our approach.

\subsection{Skeleton-based Online Clustering (SOC) Algorithm}

%\subsubsection{Overall description}

%We will now describe the Skeleton-based Online Clustering algorithm (Algorithm 1).
The key idea behind the SOC algorithm is to represent each cluster via a set of pseudorandom samples called the \textit{skeleton set}. 
The algorithm stores and constantly updates a set of 
the skeleton sets $\mathcal{P} = \{S_{1},...,S_{t}\}$. Note that the size of the set $\mathcal{P}$ corresponds to the number of clusters, 
which may change over time as new data is seen. Each skeleton set $S_{i}$
represents a cluster and consists of a sample $(v^{i}_{1},...,v^{i}_{h_{i}})$ of all the points
belonging to that cluster up to time $t$ together with some carefully chosen random numbers
$(m^{i}_{1},...,m^{i}_{h_{i}})$ and weights $(w^{i}_{1},...,w^{i}_{h_{i}})$. Thus, a skeleton set $S_{i}$ is the set of elements 
of the form: $(v^{i}_{j}, m^{i}_{j}, w^{i}_{j})$ for $j=1,2,...,h_{i}$.
Sometimes we will also call $\{v^{i}_{1},...,v^{i}_{h_{i}}\}$ a skeleton set, which should be clear from the context. 
%From the context it will be clear whether we are 
%talking about the set of data points or the corresponding set of triples.
Let us define a map $\xi(v^{i}_{j}) = m^{i}_{j}$.
We denote by $w_{y}$ the weight of the skeleton point $y$ and by $m_{y}$ the corresponding random number. 
Weights encode the local density around a skeleton point. 
We denote by $W_{i}$ the sum of weights of all the skeleton points representing the $i$th cluster
and by $W(\mathcal{A})$ the sum of weights for all the points belonging to set $\mathcal{A}$. 
The skeleton set is updated in such a way 
that at any given time $t$ skeleton points are pseudouniformly distributed in the entire cluster.
As mentioned before, the number $h_{i}$ of skeleton points of the $i$th cluster (alternative notation: $h_{S}$, 
where $S$ stands for the skeleton set) is not fixed and can vary over time.
When the cluster arises, it is initialized with $h_{init}$ skeleton points. In the algorithm we take $h_{init}=1$
and in the theoretical section we show how the lower bounds on $h_{init}$ can be translated to strict
provable guarantees regarding the quality of the produced clustering.
The algorithm tries to maintain a skeleton set in such a way that between any two skeleton points there exists
a path of relatively short edges consisting entirely of other skeleton points from the same skeleton set.
If the cluster grows and the average distance between skeleton points is too large, the number of skeleton points
is increased. This number never exceeds $H$ which is a given upper bound determining the memory the user is inclined to
delegate to encode any cluster.
%We denote by $h_{S}$ the number of skeleton points of the skeleton set $S$.

The overall algorithm works by first initializing the number of samples stored in each skeleton set. 
In this work, without loss of generality we assume that each skeleton set initially has only one sample.
We will propose two variants of the algorithm - one where only lazy cluster-merging is performed (\textit{MergeOnlySOC}) and the other, where
merging can be more aggressive since splitting clusters is also allowed (\textit{MergeSplitSOC}). As we will see in the experimental section, the latter produces a good approximation of the groundtruth clustering in fewer steps, 
but at the cost of extra time needed  to check and perform splitting.
If \textit{MergeSplitSOC} version is turned on then the algorithm keeps a set of undirected graphs $\mathcal{G}$. Each element of $\mathcal{G}$ is associated with a 
different skeleton set and encodes the topology of connections between points in that set. We denote by $G_{S}$ an element of $\mathcal{G}$ associated with the skeleton set $S$.

\begin{algorithm}[t]
%\small
%\SetAlgoLined
\textbf{Input: } Infinite data stream $\mathcal{D}$\;
\textbf{Output: } Cluster assignment for the observed data\;
Pick $r, \alpha$; 
$\mathcal{P} \leftarrow \emptyset, \mathcal{G} \leftarrow \emptyset$\;
\While{true}{
\For{each $S_{i}$} {
  \If{exists $v \in S_{i}$ such that $w_{v} \leq \frac{W_{i}}{2|S_{i}|}$} {
    $CheckSplit(\mathcal{G}, v, r, S_{i}, \mathcal{P})$\;
  }
}
Read next $x \in \mathcal{D}$\;
Compute $T_{i} = S_{i} \cap B(x,r)$ for each $S_{i} \in \mathcal{P}$\;
$\mathcal{U} \leftarrow \emptyset$\; 
\For{each $T_{i}$} {
  \If{$W(T_{i}) \geq \alpha W_{i}$}{
    $\mathcal{U} \leftarrow \mathcal{U} \cup \{S_{i}\};$ 
}
}
\eIf{$\mathcal{U} \neq \emptyset$} {
   $Merge(x, \mathcal{U}, \mathcal{P}, \mathcal{G}, r)$; 
}{
   $AddSingleton(x, \mathcal{P}, \mathcal{G})$;
   }
}
\caption{SOC clustering - main procedure}
\label{alg1}
%\vspace{-2mm}
\end{algorithm}

The overview of the SOC algorithm is given in Algorithm \ref{alg1}. If splitting is turned on, at each time step t, 
given the existing skeleton set for each cluster, the algorithm checks if any cluster should be split.  Then,  as a new data point $x$ arrives 
in the stream, first a ball of radius $r$ centered at $x$ i.e., $B(x,r)$ is created. 
Then, the intersection of this ball with each existing skeleton set is computed. If the weight of the skeleton points in this 
intersection is more than $\alpha W_{i}$ for $0 < \alpha < 1$, then $x$ is assumed to 
belong to that cluster and the corresponding skeleton set is updated. Note that it is possible that multiple clusters claim the point $x$, 
in which case, all those clusters are considered for merging. If the intersection of the ball with all the skeleton sets is empty, then a new singleton cluster is created.

We will start with describing the \textit{MergeOnlySOC} variant (i.e., assume that the 
splitting-related procedures: \textit{CheckSplit} and \textit{UpdatedGraph} in Algorithm \ref{alg1} and Algorithm \ref{alg_merge} are turned off). The \textit{MergeSplitSOC} variant will discussed later in section \ref{sec:splitting}.

%\vspace{0.3in}
\begin{algorithm}[t]
\small
\SetAlgoLined
\textbf{Input: } Datapoint $x$, subset $\mathcal{U}=\{S_{i_{1}},...,S_{i_{k}}\} \subseteq \mathcal{P}$, current clustering $\mathcal{P}$, family of graphs $\mathcal{G}$, radius $r$\;
\textbf{Output: } Updated $\mathcal{P}$ after merging $x$ with clusters from $\mathcal{G}$\;
Denote: $W = \sum_{y \in (S_{i_{1}} \cup ... \cup S_{i_{k}}) \cap B(x,r)} w_{y}$, 
$d_{av}(x,r) = \sum_{y \in (S_{i_{1}} \cup ... \cup S_{i_{k}}) \cap B(x,r)} \frac{w_{y}}{W}\|x-y\|_{2}$,\\
$h_{un} = \min(\sum_{j=1,...,k} {h_{S_{i_{j}}}}, H)$\; 
Compute: $S^{ext}_{i_{j}} = Ext(S_{i_{j}}, h_{un})$ for $j=1,...,k$\;
Let: $\mathcal{S} \leftarrow \{S^{ext}_{i_{1}},...,S^{ext}_{i_{k}}\}$\; 
\If{$d_{av}(x,r) \leq \frac{r}{2}$ or $(h_{un} = H)$} {
  $\mathcal{S} \leftarrow \mathcal{S} \cup \{Ext(\{x\},h_{un})\}$\;
}

%Generate a random sequence of $h$ numbers: $(r^{x}_{1},...,r^{x}_{h})$ according to $Gen(s)$\;
%Let $\xi^{i}(x) = r^{x}_{i}$\;
%$Sk_{merged} \leftarrow \emptyset$\;
Denote by $v^{i}_{j}$ the $jth$ skeleton point of the $ith$ skeleton set of $\mathcal{S}$\;
initialize: $S_{merged} \leftarrow \emptyset$\;
\For{$j=1,...,h_{un}$}{
$v \leftarrow argmin(\xi(v^{1}_{j}),...,\xi(v^{|\mathcal{S}|}_{j}))$;\newline
$m \leftarrow \min(\xi(v^{1}_{j}),...,\xi(v^{|\mathcal{S}|}_{j}))$;\newline
$w \leftarrow w_{v}$;\newline
$S_{merged} \leftarrow S_{merged} \cup \{(v,m, w)\}$;\newline  
}   
\eIf{$d_{av}(x,r) > \frac{r}{2}$ and $(h_{un} < H)$} {
  generate $r$ according to $Gen(s)$\;
  $S_{merged} \leftarrow S_{merged} \cup \{(x, r, 1)\}$\;
} {
 Let $v_{min} \in S$, where $S \in S_{merged}$ be such that: $\|x-v_{min}\|_{2} = \min_{S \in S_{merged}} \min_{z \in S} \|x-z\|_{2}$\;
 \If{ $v_{min} \neq x$}
  {
    $w_{v_{min}} \rightarrow w_{v_{min}}+1$\;
    $m_{v_{min}} \rightarrow \min(m_{v_{min}}, r)$, where $r$ is generated according to $Gen(s)$\;
  }
}
Let $\mathcal{G}_{0} = \{G_{S_{i_{1}}},...,G_{S_{i_{k}}}\}$\;
$\mathcal{P} \leftarrow (\mathcal{P} \cup \{S_{merged}\}) \backslash \{S_{i_{1}},...,S_{i_{k}}\};$\newline
$UpdatedGraph(\mathcal{G}, x, r, \mathcal{G}_{0}, S_{merged})$\;
\caption{SOC clustering - \textit{Merge} subprocedure}
\label{alg_merge}
\end{algorithm}

\subsubsection{Subprocedure \textit{Merge}}

The goal of the \textit{Merge} subprocedure is to merge a new point $x$ with one or more clusters (Algorithm \ref{alg_merge}). 
When $x$ is assigned to multiple clusters, 
it basically acts as a linking point to merge them, resulting in a unified cluster. 
This step is crucial in online clustering scenario as points from the same cluster 
may be assigned to different clusters initially but as more evidence builds up from the new data, one can combine these clusters to recover the true underlying cluster. 
%As we show later in theoretical analysis (Sec. \ref{sec:theory}), the proposed algorithm with high probability does not merge points from different clusters. 
In the \textit{Merge} subprocedure the skeleton set is updated when a new cluster is constructed.
The skeleton sets $S_{i_{1}},...,S_{i_{k}}$ representing clusters that will be merged are given as input.
Before describing the \textit{Merge} procedure let us introduce an important subroutine.
Let $S=\{(v_{1},m_{1},w_{1}),...,(v_{t},m_{t},w_{t})\}$ be a skeleton set of size $h_{S} < h$ for some $h>0$.
We denote by $Ext(S, h)$ the extension of $S$ obtained by adding to $S$ exactly $h-h_{S}$ more triples according to the following procedure.
Each newly added triple has weight $1$. Each newly added skeleton point is chosen independently at random from the set $\{v_{1},...,v_{t}\}$
in such a way that skeleton point $v_{i}$ is chosen with probability $\frac{w_{i}}{\sum_{j=1}^{t} w_{j}}$.
The corresponding random number $m$ is generated by a 
pseudorandom number generator $Gen(s)$ with seed $s$. The seed can be initialized randomly or alternatively 
it can be chosen as a function of the skeleton point by conceptually partitioning the entire input space with a grid of lengh $\delta$, 
and using the id of the cell occupied by $x$ in this grid. The latter procedure is useful for infinite streams to avoid correlated 
random sequences for far away points in the space. 
Subroutine $Ext$ can be run also if its first argument is a single point $x$. In this case the output is a skeleton set of the form
$\{(x,r_{1},1),...,(x,r_{h},1)\}$, where $r_{1},...,r_{h}$ is a sequence of $h$ random numbers generated according to $Gen(s)$.

The \textit{Merge} algorithm computes an average weighted distance $d_{av}(x,r)$ from the new point to those skeleton points from $S_{i_{1}},...,S_{i_{k}}$
that reside in $B(x,r)$.
Next, two cases are considered.
If $d_{av}(x,r) \leq \frac{r}{2}$ or the number of skeleton points in all the skeleton sets to be merged is $H$ 
then the algorithm decides not to increase the size of the merged skeleton set (since the linking point is relatively close to the merged clusters 
or the union of skeleton sets under consideration is already saturated).
Denote by $h_{un}$ the minimum of $H$ and the total number of skeleton points in all skeleton sets to be merged. The merged skeleton set will be of size $h_{un}$.
Let us describe how the $jth$ skeleton point of the newly formed cluster for $j=1,...,h_{un}$ will be computed.
First, each contributing skeleton set is extended to size $h_{un}$ by weighted random sampling from it (see procedure $Ext$ described above). 
This is also the case for $x$ (we treat $x$ as a skeleton set consisting of $h_{un}$ copies of $x$). 

We take the $j^{th}$ skeleton points from all the clusters to be merged and $x$, and choose the new point and the corresponding value as shown in Algorithm (\ref{alg_merge}),
using random sequence generated for $x$. 
Each newly added point has to contribute to the weight distribution in the cluster. If point $x$ is not in the new skeleton set then
the closest point $v_{min}$ from the skeleton set is found and its weight is increased by one ($x$ contributes to the total weight of $v_{min}$). 
The new skeleton set replaces the skeleton sets of all the clusters that are merged.

Now let us assume that $d_{av}(x,r) > \frac{r}{2}$ and the total number of all skeleton points in the skeleton sets to be merged is smaller than $H$.
Intuitivaly speaking, this means that the cluster had grown too much (the local density within linking point $x$ is too small)
and thus the number of skeleton points encoding the cluster has to increase (since the pool of the skeleton points that can be used to represent a cluster has not been used entirely).
If this is the case then the same procedure as for the previous case is conducted but the skeleton set corresponding to $x$ is excluded.
Finally, $x$ is added as the last skeleton-singleton of weight $1$ (and with corresponding random number selected according to a 
given random number generator)  to the newly formed cluster.

%Notice that from the main SOC procedure we know that in the ball $B(x,r)$ at least a fraction $\alpha$ of the total weight of all the skeleton points
%from the skeleton sets $Sk_{i_{1}},...,Sk_{i_{k}}$ resides. If this is the case also for the ball $B(x,\frac{r}{2})$

\subsubsection{Subprocedure \textit{AddSingleton}}

The procedure \textit{AddSingleton} adds a new cluster consisting of just $x$ when no existing cluster is found to be close enough to $x$ based on the intersection 
with the skeleton sets (Algorithm \ref{alg_singleton}). 
Next, a skeleton set for this new singleton cluster is created. 
Since a skeleton set aims 
to cover uniformly the entire mass of the cluster 
using $h_{init}$ random samples, point $x$ is repeated $h_{init}$ times to form the skeleton set for the cluster-singleton. \newline

Furthermore, a sequence of $h_{init}$ random values 
is generated from $Gen(s)$, and each copy of $x$ in the skeleton set is assigned one of the values and weight $w=1$ to build the $h_{init}$ triples and complete the skeleton.
In the proposed implementation we initialize each newly created cluster-singleton with only one skeleton point, thus $h_{init}=1$ (see Algorithm \ref{alg_singleton}).
For the newly created skeleton set an undirected graph-singleton $G_{sin}$ is created.

%\vspace*{-0.8cm}
\begin{algorithm}[t]
\small
\SetAlgoLined
\textbf{Input: } Datapoint $x$, current clustering $\mathcal{P}$, family of graphs $\mathcal{G}$\;
\textbf{Output: } Updated version of $\mathcal{P}$ after adding cluster-singleton $\{x\}$\;
Generate a random number $r^{x}$ according to $Gen(s)$\;
$S_{new} \leftarrow \{(x,r^{x},1)\}$\;
%\For{$j=1,...,h$} {
%  $Sk_{new} \leftarrow Sk_{new} \cup \{(x, r^{x}_{j})\}$;
%}
$\mathcal{P} \leftarrow \mathcal{P} \cup \{S_{new}\}$\;
$\mathcal{G} \leftarrow \mathcal{G} \cup \{G_{sin}\}$\;
\caption{SOC clustering - \textit{AddSingleton} subprocedure}
\label{alg_singleton}
\end{algorithm}
%\vspace*{-0.3cm}

\subsection{Splitting clusters}
\label{sec:splitting}
We now describe the cluster splitting procedure in the \textit{MergeSplitSOC} variant of the algorithm.
It is handled by two additional procedures: \textit{CheckSplit}
and \textit{UpdatedGraph}.

\textit{CheckSplit} determines whether a given skeleton set should be split by looking for a breaking point $v$, which is the skeleton point whose weight is at most half of the average weight of the points within the skeleton set. 
If such a point $v$ is found then the algorithm determines whether the cluster should be split as follows. First, all the points from $B(v,\frac{r}{2})$ are deleted from the corresponding graph $G$ of the skeleton set. 
A connected component analysis is then conducted on the remaining graph. If more than one connected components is found, it means $v$ is a breaking point and the cluster is split so that each connected component forms a new cluster and the points in the connected components consitute the new skeleton sets. 

The \textit{UpdatedGraph} procedure is shown in Algorithm~\ref{alg_updategraph}, and is responsible for constructing a graph $G_{union}$ for the newly formed cluster
and replacing with it all graphs corresponding to merged skeleton sets.
The graph $G_{union}$ is constructed by combining all elements of $\mathcal{G}$ corresponding to the skeleton sets that need to be merged.
Those graphs are combined by adding edges between skeleton points from the newly constructed skeleton set that are in the close neighborhood of the linking point $x$.  
In the description of \textit{UpdatedGraph} we denote by $G = G_{1} \otimes G_{2} ... \otimes G_{k}$ the graph with vertex set $V(G) = V(G_{1}) \cup ... \cup V(G_{k})$ and edge
set $E(G) = E(G_{1}) \cup ... \cup E(G_{k})$.

\begin{algorithm}[t]
\small
\SetAlgoLined
\textbf{Input: } Family of graphs $\mathcal{G}$, skeleton point $v$, radius $r$, skeleton set $S$ corresponding to $v$, current clustering $\mathcal{P}$\;
\textbf{Output: } Updated version of $\mathcal{G}$ and $\mathcal{P}$\;
Let $Rem = \{y \in S: \|v-y\|_{2} \leq \frac{r}{2}\}$\;
Let $G^{del} = G_{S} \setminus Rem$\;
Run CC algorithm on $G^{del}$ to obtain $\{\mathcal{C}_{1},...,\mathcal{C}_{t}\}$\;  
\If{$t>1$} {
  Denote by $S_{i}$ the subset of $S$ corresponding to $\mathcal{C}_{i}$ for $i=1,...,t$\;
  Update: $\mathcal{G} \leftarrow (\mathcal{G} \cup \{\mathcal{C}_{1},...,\mathcal{C}_{t}\}) \setminus \{G_{S}\}$ and\
          $\mathcal{P} \leftarrow (\mathcal{P} \cup \{S_{1},...,S_{t}\} \setminus \{S\})$\;
}

\caption{SOC clustering - \textit{CheckSplit} subroutine}
\label{alg_checksplit}
\end{algorithm}

\begin{algorithm}[t]
\small
\SetAlgoLined
\textbf{Input: } Family of graphs $\mathcal{G}$, skeleton point $x$, radius $r$,
                 subfamily $\{G_{S_{1}},...,G_{S_{k}}\} \subseteq \mathcal{G}$,
                 skeleton set $S_{new}$\;
\textbf{Output: } Updated version of $\mathcal{G}$\;
Let $G = G_{S_{1}} \otimes ...\otimes G_{S_{k}}$\;
Let $E=\{\{x,y\}: x,y \in S_{new} \cap B(x,r), \exists_{i \neq j} x \in G_{S_{i}}, y \in G_{S_{j}}\}$\;
Let $G_{merged}$ be a graph obtained from $G$ by adding edges from $E$\;
Update: $\mathcal{G} \leftarrow (\mathcal{G} \cup \{G_{merged}\}) \setminus \{G_{S_{1}},...,G_{S_{k}}\}$\;
\caption{SOC clustering - \textit{UpdatedGraph} subroutine}
\label{alg_updategraph}
\end{algorithm}

\section{Theoretical analysis}
\label{sec:theory}

In this section we provide theoretical results regarding SOC algorithm for the clustering model described in Sec. \ref{sec:clustering}. 
We start by introducing the general mathematical model we are about to analyze. It is one of the many variants of planted partition models
used to construct data with hard clustering and outliers. Notice that our algorithm does not require the input to be produced according to this model.
In particular, we do not use any specific parameters of the model in the algorithm.

For a set of $D$-dimensional data, we assume it contains $k$ disjoint compact sets , which are called \textit{cores} and denoted as $\mathcal{R}_{1},...,\mathcal{R}_{k} \subseteq \mathbb{R}^{D}$. The cores are  called $\Delta$-separable if the minimum distance between any two cores is greater than $\Delta$, i.e.: 
$\forall_{1\leq i < j \leq k, x \in \mathcal{R}_{i}, y \in \mathcal{R}_{j}} \|x-y\|_{2} > \Delta$. 
These cores can have arbitrary shapes giving rise to the observed clusters such that the points in the $i^{th}$ cluster $\mathcal{C}_{i}$ 
come from core $\mathcal{R}_i$ with high probability and from the rest of the space with low probability. 
Formally, given a set of probabilities $\{p_{1},...p_{k}\}$, points in the cluster $\mathcal{C}_{i}$ are sampled from the core $\mathcal{R}_i$ with probability $p_i$ and from outside $\mathcal{R}_i$ with probability $1-p_i$, where $p_i \gg (1-p_i)$. 
It is important to note that even though the cores are separable, this is no longer the case for clusters due to the presence of "outliers". 
In other words, short-edge paths between points from different clusters may exist, but not many in expectation.
We call the clustering model presented above as a \textit{$(\tilde{\mathcal{R}},\tilde{p})$-model}, where $\tilde{p}=(p_{1},...,p_{k})$ and $\tilde{\mathcal{R}} = (\mathcal{R}_{1},...,\mathcal{R}_{k})$.
This is a quite general model which has good-quality clustering of cores 
(because of $\Delta$-separability). However due to the outliers, the task of recovering the clusters is 
nontrivial even in an offline setting. Simple heuristics such as connected-components cannot be used to recover the clusters in the offline mode. 
The online setting brings additional algorithmic and computational challenges. Below, we give details of the proposed clustering mechanism. 

We need the following definition.

\begin{definition}
A set $\mathcal{W} \subseteq R^{D}$
is called to be \textit{$(s,r)$-coverable} if $\mathcal{W}$ can be covered by $s$ balls, each of radius r.
\end{definition}

Let $\pi_{i}$ be the probability that a new point in the data stream belongs to cluster $\mathcal{C}_{i}$. Fix a covering $\mathcal{C}$ of $\mathcal{R}_{i}$ with $|\mathcal{C}| \le s$ (for every core $\mathcal{R}_{i}$).
Let $\mathcal{B} \in \mathcal{C}$ be an arbitrary ball from the covering $\mathcal{C}$.
Furthermore, let $p^{b}_{i}$ be a lower bound on the probability that a new point is  
from set ($\mathcal{R}_{i} \cap \mathcal{B}$) given it belongs to core $\mathcal{R}_{i}$, which can be expressed as $p^{b}_{i} \sim \frac{1}{s}$.
Denote $\Gamma_{i} = \frac{\pi_{i}p_{i}p_{i}^{b}}{p^{f} + \pi_{i}p_{i}}$, 
and $\gamma_{i} = \frac{p^{f}}{p^{f} + \pi_{i}p_{i}}$, where
$p^{f} = \sum_{i=1}^{k}p_{i}^{f}$, and $p_i^f = 1 - p_i$.
Here $\Gamma_{i}$ is the probability that a new point came from a fixed
ball of covering $\mathcal{C}$ of $\mathcal{R}_{i}$ given that it belongs to cluster $\mathcal{C}_{i}$. Similarly, $\gamma_{i}$ is the probabiltiy that a new point is an outlier
given that it belongs to cluster $\mathcal{C}_{i}$.
Since outliers are expected to be lower than the points from the cluster cores, $\gamma_{i} \ll \Gamma_{i}$. Denote $\Gamma = \min_{i} \Gamma_{i}$ and $\gamma = \max_{i} \gamma_{i}$.

Since we keep at most $H$ samples in a skeleton set, most of the cluster points are not in this set.
%Whenever we say that a data point belongs to the cluster we mean that at some point of the algorithm
%this point was added to that cluster (but not necessarily became its skeleton point).
The error made by the algorithm on a new point $v$ is defined as follows: Suppose $v$ comes from the 
core $\mathcal{R}_i$, and gets assigned to a cluster that contains points from other cores as well, 
or there exists another cluster that also contains points from $\mathcal{R}_i$. 
Note that it is a strict definition of error as in an online setting transient overclustering is expected due to lack of enough data in the early phase. 
We say that the algorithm reaches the \textit{saturation phase} when each skeleton set reaches size $H$.
We are ready to state the main results of our analysis regarding the \textit{MergeOnlySOC} version of the algorithm.

%In particular, if there is one big cluster constructed by the algorithm with many points from
%the same core as $v$ but there is at least one other cluster containing some points of that
%groundtruth core, we say that the algorithm makes a mistake on $v$. Surprisingly, even under
%this strict definition (that in practice can be significantly relaxed), we can give some
%theoretical guarantees as we show below.

\begin{theorem}
\label{wrongmergetheorem}
Assume that we are given a dataset constructed according to the $(\tilde{\mathcal{R}},\tilde{p})$-model with 
$k$ cores with outliers. Cores are $\Delta$-separable. Assume that each core of the cluster is $(s,\frac{\Delta}{4})$-coverable.
Let $n$ be the number of all the points seen by the algorithm after the saturation phase has been reached. 
Then with probability at least $1 - 4\epsilon$ for 
$h_{init} = \Omega(\max(\frac{\gamma}{\Gamma^{3}}\log(\frac{sk}{\epsilon}), \frac{1}{\Gamma^{2}}\log(\frac{nsk}{\epsilon})))$,
$r = \frac{\Delta}{4}$ and $\alpha = \frac{\Gamma}{4}$ the SOC algorithm will not merge clusters containing points from different cores 
in the saturation phase if they were not merged earlier.
\end{theorem}

Theorem \ref{wrongmergetheorem} gives upper bounds on the minimal number of skeleton points per cluster ensuring that $\textit{MergeOnlySOC}$
does not undercluster. As we will see in the experimental section, this bound in practice is much lower.
We also have the following.
 
\begin{theorem}
\label{rightmergetheorem}
%Fix some core $\mathcal{R}_{i}$.
Under the assumptions from Theorem \ref{wrongmergetheorem}, with probability at least $1 - \epsilon$, the SOC algorithm
will not make any errors on points coming from core $\mathcal{R}_{i}$ after $m = \frac{17}{2\Gamma^{2}}\log(\frac{3s}{\epsilon})$ 
points from the corresponding cluster $\mathcal{C}_{i}$ have been seen in the saturation phase.
\end{theorem}

Theorem \ref{rightmergetheorem} says that under a reasonable assumption regarding
the number of initial skeleton points per cluster and after the short initial subphase of the saturation phase,  
algorithm $\textit{MergeOnlySOC}$ classifies correctly all points coming from cores. In other words, we obtain
the upper bound on the rate of convergence of the number of clusters produced by $\textit{MergeOnlySOC}$ to the groundtruth value.

The proofs of Theorem \ref{wrongmergetheorem} and Theorem \ref{rightmergetheorem}  will be given in the Appendix. 
Below we give a very short introduction and present a useful lemma that we will rely on later.

\begin{figure}
  \center
\includegraphics[width = 3.3in]{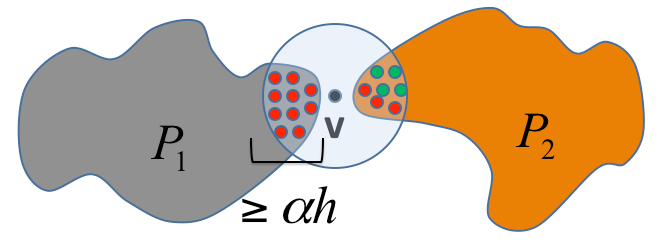} 
\caption{Merge scenario as in Lemma \ref{mergelemma}. Data point $v$ merges two clusters: $P_{1}$ and $P_{2}$.
The intersection $\mathcal{B}(v,\frac{\Delta}{2} \cap P_{i})$ for some $i \in \{1,2\}$ must consist
only of outliers (points marked red). The other intersection may contain points from the core (points marked green).}
\label{fig:seven}
\end{figure}

\subsection{Merging Lemma}

Since both theorems consider the saturation phase of the algorithm, in the theoretical analysis whenever we will talk about the
algorithm we will in fact mean its saturation phase.
Without loss of generality we can also assume in our theoretical analysis that each skeleton point has weight one.
Indeed, a point $v$ that has weight $w_{v}>0$ may be equivalently treated as a collection of $w_{v}$ skeleton points
of weight one (see: description of the algorithm).
Let us formulate the following lemma:

\begin{lemma}
 \label{mergelemma}
 Let us assume that at time $t$ the algorithm merges 
 two clusters: $P_1$ and $P_2$ 
 such that $P_1$ contains a point from $\mathcal{R}_{i}$
 and $P_{2}$ contains a point from $\mathcal{R}_{j}$ for some $i \neq j$.
 Then either: at least $\alpha h$ of all skeleton points of $P_{1}$ at time $t$ are outliers
 or: at least $\alpha h$ of all skeleton points of $P_{2}$ at time $t$ are outliers.
\end{lemma}

\begin{figure*}[t]
	\centering
	\includegraphics[width=\textwidth, height=7.5cm]{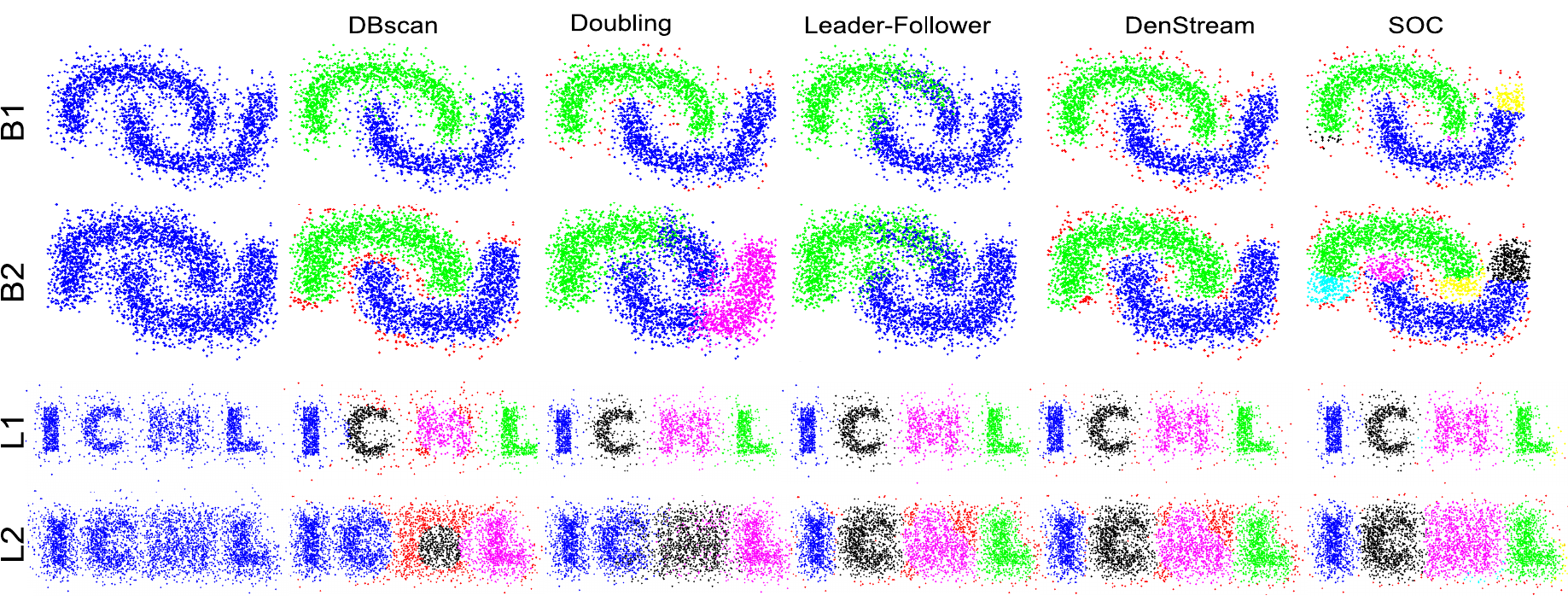}
	\caption{Clustering results for the different methods. The left column shows four different synthetic datasets. 
	         The resulting clusters for each method are marked using different colors. Note that DenStream is a hybrid online-offline technique while
	         SOC is a purely online method.}
	\label{fig:syn_results}
\end{figure*}

\begin{proof}
The lemma follows from the definition of $\Delta$, according to which, any two points taken from different
cores are at least $\Delta$ distance apart.
If two clusters: $P_{1}$, $P_{2}$ are merged at time $t$ then there exists a data point $v$ (a merger) and a 
ball $\mathcal{B}(v,\frac{\Delta}{2})$ such that $\mathcal{B}(v,\frac{\Delta}{2})$ contains at least $\alpha h$ skeleton points of
$P_{1}$ and at least $\alpha h$ skeleton points of $P_{2}$ (see: Fig. \ref{fig:seven}). But $\mathcal{B}(v,\frac{\Delta}{2})$ cannot contain points from different cores since they
are $\Delta$-separable. Thus at least one of the clusters from: $\{P_{1},P_{2}\}$
has in $\mathcal{B}(v,\frac{\Delta}{2})$ at least $\alpha h$ skeleton points that are outliers.
\end{proof}

\section{Experiments}

%The characteristic of the nonparametric clustering setting is that, in contrary to the parametric one, there does not exist a simple function that needs to be optimized and 
%that can be given by a compact-form expression. Thus, almost all the nonparametric clustering works measure the quality of the clustering via manual inspection.
%The other measure is clusters' purity, but this one is inherently entangled with the number of clusters (one can always ouput the partitioning of all the data 
%into clusters-singletons).

We evaluated the performance of the proposed SOC algorithm using four synthetic 
datasets as shown in  Fig.~\ref{fig:syn_results} (left column).
The sets contain data points in 20 dimensions.  
The first two dimensions were randomly drawn from predefined clusters, as shown in the figure, while the other 18 dimensions was random noise. 
For the data sets B1 and B2, 1000 data points were randomly drawn from each of the two banana shaped clusters for the first two dimensions, 
Then 1000 (for B1) and 2000 (for B2) outliers were randomly generated from a vicinity of the shapes, respectively. 
For the data sets L1 and L2, 500 data points were randomly drawn from each of the four letter shaped clusters.
Then 500 (for L1) and 2500 (for L2) outliers were randomly generated  from a vicinity of the shapes respectively. The values in the other 18 dimensions for all the data were Gaussian white noise with a standard deviation of 0.01. All the data points were then randomly permuted so that their orders in the data stream would not affect the results. 
Examples of the datasets were plotted in the first two dimensions and shown in Fig.~\ref{fig:syn_results} (left column). 
We used the \textit{MergeSplitSOC} version of the algorithm since it provided faster convergence of the number of clusters under the same quality guarantees.

We compared the SOC method with several state-of-the-art nonparametric clustering methods, i.e.,  
DBScan~\cite{ester}, Leader-Follower algorithm~\cite{duda}, Doubling algorithm~\cite{charikar}, and DenStream~\cite{cao}. 
The clustering quality was quantitatively evaluated using the average clustering purity, which is defined as
\begin{equation}
	p = \frac{\sum_{i=1}^K\frac{|C_i^d|}{|C_i|}}{K} \,,
\end{equation}
where $K$ is the number of clusters, $|C_i|$is the number of points in cluster $i$, and $|C_i^d|$ is the number of points in cluster $i$ with the dominant class label.

\begin{figure}[t]
	\centering
	\includegraphics[width=8cm]{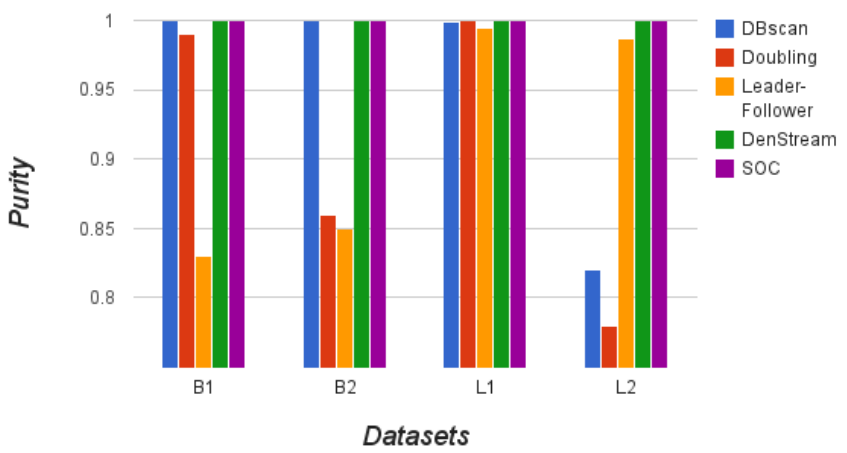}
	\caption{The clustering purity for different methods on the four synthetic data sets. }
	\label{fig:syn_purity}
\end{figure}

Fig.~\ref{fig:syn_results} shows the comparative results of the SoC method as well as several other methods.
Fig.~\ref{fig:syn_purity} shows the clustering purity of all the methods. 
The SOC method requires two parameters ($\alpha$ and $r$). In all the experiments, we selected $\alpha = 0.03$ and $r=0.07$. 
All the results were produced using the best choice of parameters for each method. We note that parameter tuning was not trivial because most of the methods require at least two parameters. 

Results showed that the SOC method was able to cluster the data well, even though it slightly over-clustered in the datasets B1 and B2. 
The Leader-Follower algorithm as well as streaming DBScan simply do not handle this type of data.
SOC obtains similar results to these from DenStream algorithm (it produced slight overclustering, but obtained almost 100\% purity). 
DBscan worked well on the banana sets, but it failed to cluster in L2, where the outliers over-numbered the true clusters. 
Doubling failed to work on the more noisy data sets (B2 and L2). For the other two datasets, the clustering purity was low, probably because of the noise in other 18 dimensions. 
Leader-Follower method worked fine for L1 and L2, but poorly for B1 and B2. 
It was mostly because of the nature of the method, and partly because of the noise in the other 18 dimensions.
Standard variant of the Leader-Follower uses only a small number of centers per cluster and when the clusters are not contained
in convex well-separable objects, the recognition is very poor.
DenStream worked well on all the cases. 

\begin{figure}[t]
\centering
\begin{tabular}{cc}
	\includegraphics[width=5cm]{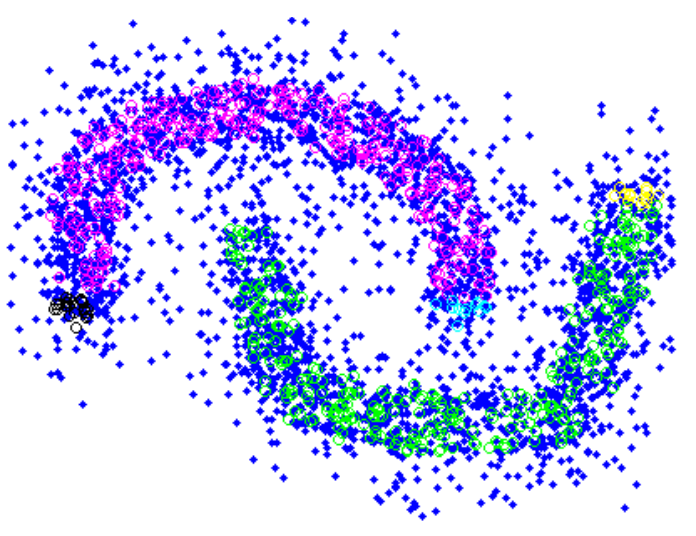} \\ 
	(a) B1\\
	\includegraphics[width=6cm]{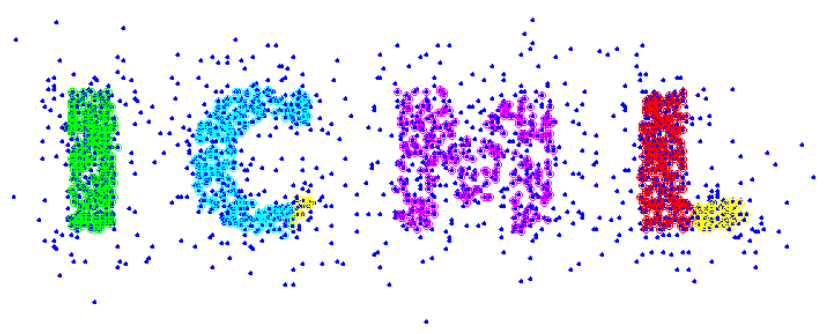} \\
	 (b) L1
\end{tabular}

\caption{The skeleton points for the data sets (a) B1 and (b) L1. The data points are colored blue, and the skeleton points for different clusters are marked with different colors. }
\label{fig:syn_skeletons}
\end{figure}

Our method is faster than DenStream (the running times per point varied from $60$ to $90$ microseconds for SOC and were above $100$ microseconds for DenStream).
The reason is that DenStream is not a purely online approach and performs offline clustering periodically. That part is a bit expensive.
DenStream has another serious drawback. The id of the cluster the newly coming point is assigned to is computed based on the most recent snapshot of the 
offline clustering that was produced, not on-fly. Thus the accuracy of the method depends heavily on the special parameter determining 
how much data distribution evolves over time.
Since parameter tuning is always very nontrivial for the density-based methods, this one extra parameter adds another layer of difficulty.
We do not need this parameter in our approach. 

The SOC method slightly overclustered in B1 and B2 because of the online nature of the algorithm and the presence of the outliers. 
Nontheless, it was able to correctly throw out the outliers and produced results with high purity.  
This is because, even though outliers can become part of the skeleton set of a cluster, they are 
typically replaced by the true cluster points eventually as the true cluster points have a higher density and arrive at a higher rate than  the outliers. 

Fig.~\ref{fig:syn_skeletons} shows the skeleton points generated by SOC for two data sets. The maximum number of skeleton
points used was only few hundred points. The number of skeleton points grows gradually as clusters become bigger.
We set up the upper bound on the skeleton points per cluster as $H=400$ but this number was not reached.
\section{Conclusions}

We have presented a new truly online clustering algorithm which can recover arbitrary-shaped clusters in the 
presence of outliers from massive data streams. Each cluster is represented efficiently by a skeleton set, 
which is continuously updated to dynamically adapt to the changing data distribution. The proposed technique is 
theoretically sound as well as fast and space-efficient in practice. It produced good-quality clusters in various experiments
for nonconvex clusters. It outperforms several online approaches on many datasets and produces results similar to the most effective hybrid ones
that combine online and offline steps (such as DenStream).
In the future, we would like to investigate other methods for updating skeletons within given framework in the online fashion since this mechanism is crucial for the
effectiveness of the presented approach.
The other interesting area is the research on the maximal number of clusters that are created during the execution of the algorithm.
A more precise bound will provide a more accurate theoretical estimate on the memory usage.

\bibliography{fast_online_clustering}

\begin{thebibliography}{22}
\providecommand{\natexlab}[1]{#1}
\providecommand{\url}[1]{\texttt{#1}}
\expandafter\ifx\csname urlstyle\endcsname\relax
  \providecommand{\doi}[1]{doi: #1}\else
  \providecommand{\doi}{doi: \begingroup \urlstyle{rm}\Url}\fi

\bibitem[Aggarwal et~al.(2003)Aggarwal, Han, Wang, and Yu]{aggarwal}
Aggarwal, Charu~C., Han, Jiawei, Wang, Jianyong, and Yu, Philip~S.
\newblock A framework for clustering evolving data streams.
\newblock In \emph{{VLDB}}, pp.\  81--92, 2003.

\bibitem[Ailon et~al.(2009)Ailon, Jaiswal, and Monteleoni]{ailon}
Ailon, N., Jaiswal, R., and Monteleoni, C.
\newblock Streaming k-means approximation.
\newblock \emph{Neural Processing Systems Conference}, pp.\  10--18, 2009.

\bibitem[Alon et~al.(2000)Alon, Dar, Parnas, and Ron]{alon}
Alon, Noga, Dar, Seannie, Parnas, Michal, and Ron, Dana.
\newblock Testing of clustering.
\newblock In \emph{41st Annual Symposium on Foundations of Computer Science,
  {FOCS} 2000, 12-14 November 2000, Redondo Beach, California, {USA}}, pp.\
  240--250, 2000.

\bibitem[Amini et~al.(2014)Amini, Teh, and Saboohi]{amini}
Amini, Amineh, Teh, Ying~Wah, and Saboohi, Hadi.
\newblock On density-based data streams clustering algorithms: {A} survey.
\newblock \emph{J. Comput. Sci. Technol.}, 29\penalty0 (1):\penalty0 116--141,
  2014.

\bibitem[B\={a}doiu et~al.(2002)B\={a}doiu, Har-Peled, and Indyk]{indyk}
B\={a}doiu, Mihai, Har-Peled, Sariel, and Indyk, Piotr.
\newblock Approximate clustering via core-sets.
\newblock In \emph{Proceedings of the Thiry-fourth Annual ACM Symposium on
  Theory of Computing}, STOC '02, pp.\  250--257. ACM, 2002.

\bibitem[Bagirov et~al.(2011)Bagirov, Ugon, and Webb]{bagirov}
Bagirov, A., Ugon, J., and Webb, D.
\newblock Fast modified global k-means algorithm for incremental cluster
  recognition.
\newblock \emph{Pattern Recognition}, 44:\penalty0 866--876, 2011.

\bibitem[Cao et~al.(2006)Cao, Ester, Qian, and Zhou]{cao}
Cao, Feng, Ester, Martin, Qian, Weining, and Zhou, Aoying.
\newblock Density-based clustering over an evolving data stream with noise.
\newblock In \emph{Proceedings of the Sixth {SIAM} International Conference on
  Data Mining, April 20-22, 2006, Bethesda, MD, {USA}}, pp.\  328--339, 2006.

\bibitem[Charikar et~al.(1997)Charikar, Chekuri, Feder, and Motwani]{charikar}
Charikar, Moses, Chekuri, Chandra, Feder, Tom\'{a}s, and Motwani, Rajeev.
\newblock Incremental clustering and dynamic information retrieval.
\newblock In \emph{Proceedings of the Twenty-ninth Annual ACM Symposium on
  Theory of Computing}, STOC '97, pp.\  626--635. ACM, 1997.

\bibitem[Chia et~al.(2009)Chia, Song, Zhou, Hino, and Tseng]{song}
Chia, Y., Song, X., Zhou, D., Hino, K., and Tseng, B.
\newblock On evolutionary spectral clustering.
\newblock \emph{TKDD}, 3, 2009.

\bibitem[de~Andrade~Silva et~al.(2013)de~Andrade~Silva, Faria, Barros,
  Hruschka, de~Carvalho, and Gama]{silva}
de~Andrade~Silva, Jonathan, Faria, Elaine~R., Barros, Rodrigo~C., Hruschka,
  Eduardo~R., de~Carvalho, Andr{\'{e}} Carlos Ponce Leon~Ferreira, and Gama,
  Jo{\~{a}}o.
\newblock Data stream clustering: {A} survey.
\newblock \emph{{ACM} Comput. Surv.}, 46\penalty0 (1):\penalty0 13, 2013.

\bibitem[Duda et~al.(2000)Duda, Hart, and Stork]{duda}
Duda, Richard~O., Hart, Peter~E., and Stork, David~G.
\newblock \emph{Pattern Classification (2Nd Edition)}.
\newblock Wiley-Interscience, 2000.

\bibitem[Ester et~al.(1996)Ester, peter Kriegel, S, and Xu]{ester}
Ester, Martin, peter Kriegel, Hans, S, Jörg, and Xu, Xiaowei.
\newblock A density-based algorithm for discovering clusters in large spatial
  databases with noise.
\newblock pp.\  226--231. AAAI Press, 1996.

\bibitem[Gonzalez(1985)]{gonzalez}
Gonzalez, Teofilo~F.
\newblock Clustering to minimize the maximum intercluster distance.
\newblock \emph{Theor. Comput. Sci.}, 38:\penalty0 293--306, 1985.

\bibitem[Guha et~al.(2003)Guha, Meyerson, Mishra, Motwani, and
  O'Callaghan]{guha}
Guha, S., Meyerson, A., Mishra, N., Motwani, R., and O'Callaghan, L.
\newblock Clustering data streams: Theory and practice.
\newblock \emph{IEEE Transactions on Knowledge and Data Engineering},
  15:\penalty0 515--528, 2003.

\bibitem[Guha et~al.(2001)Guha, Rastogi, and Shim]{rastogi}
Guha, Sudipto, Rastogi, Rajeev, and Shim, Kyuseok.
\newblock Cure: An efficient clustering algorithm for large databases.
\newblock \emph{Inf. Syst.}, 26\penalty0 (1):\penalty0 35--58, 2001.

\bibitem[Har-peled \& Varadarajan(2001)Har-peled and Varadarajan]{varadarajan}
Har-peled, Sariel and Varadarajan, Kasturi~R.
\newblock Approximate shape fitting via linearization.
\newblock In \emph{In Proc. 42nd Annu. IEEE Sympos. Found. Comput. Sci}, pp.\
  66--73, 2001.

\bibitem[Jia(2012)]{jia}
Jia, Y.
\newblock Online spectral clustering on network streams.
\newblock \emph{Dissertation at the Department of Electrical Engineering and
  Computer Science, University of Kansas}, 2012.

\bibitem[Langone et~al.(2014)Langone, Agudelo, Moor, and Suykens]{langone}
Langone, R., Agudelo, O., Moor, B.~De, and Suykens, J.
\newblock Incremental kernel spectral clustering for online learning of
  non-stationary data.
\newblock \emph{Neurocomputing}, 2014.

\bibitem[Ng et~al.(2001)Ng, Jordan, and Weiss]{ng}
Ng, Andrew~Y., Jordan, Michael~I., and Weiss, Yair.
\newblock On spectral clustering: Analysis and an algorithm.
\newblock In \emph{Advances in Neural Information Processing Systems 14 [Neural
  Information Processing Systems: Natural and Synthetic, {NIPS} 2001, December
  3-8, 2001, Vancouver, British Columbia, Canada]}, pp.\  849--856, 2001.

\bibitem[Ning et~al.(2010)Ning, Xu, Chi, Gong, and Huang]{ning}
Ning, H., Xu, W., Chi, Y., Gong, Y., and Huang, T.
\newblock Incremental spectral clustering by efficiently updating the
  eigen-system.
\newblock \emph{Pattern Recognition}, pp.\  113--127, 2010.

\bibitem[Shah \& Zaman(2010)Shah and Zaman]{shah}
Shah, Devavrat and Zaman, Tauhid.
\newblock Community detection in networks: The leader-follower algorithm.
\newblock \emph{CoRR}, abs/1011.0774, 2010.

\bibitem[Shindler et~al.(2011)Shindler, Wong, and Meyerson]{meyerson}
Shindler, M., Wong, A., and Meyerson, A.
\newblock Fast and accurate k-means for large datasets.
\newblock \emph{Advances in Neural Information Processing Systems 24: 25th
  Annual Conference on Neural Information Processing Systems Conference}, pp.\
  2375--2383, 2011.

\end{thebibliography}
\bibliographystyle{icml2015}

%\newpage

\section{Appendix}

\subsection{Proof of Theorem \ref{wrongmergetheorem}}

\begin{proof}
From Lemma \ref{mergelemma} we conclude that in order to find an upper bound on the event
that the algorithm will at some point merge "wrong clusters", it suffices to find the upper bound on the
probability that the algorithm will at some point produce a cluster with at least $\alpha h$ skeleton points
that are outliers and at least one nonoutlier $v$. Denote this latter event by $\mathcal{E}$.
Denote by $\mathcal{E}_{i}$ the intersection of $\mathcal{E}$ with the event that $v$ comes from the
core $\mathcal{R}_{i}$. Note that $\mathcal{E} = \mathcal{E}_{1}+...+\mathcal{E}_{k}$.

%We first give the proof sketch. Let $\mathcal{C}_{i}$ be a fixed cluster.
%In the first phase of the algorithm we want to collect several skeleton points from every ball of the covering of $\mathcal{R}_{i}$. It should be done before at least $\frac{\alpha h}{2}$ outliers become  
%a part of the skeleton set. We want to upper bound the probability that at least $\frac{\alpha h}{2}$ outliers become part of a skeleton set. This is what is done in the formal proof below.
%
%We want to say that this will hold with very large probability. If fact we need a stronger
%result, we want this to be true from that time on.
%In a nutshell, we want to say that the skeleton of $\mathcal{R}_{i}$ will grow fast enough.
%When this is the case then we analyze the second phase of the algorithm.
%In this phase we again focus on the analysis of the chance that the new arriving outlier becomes the part of the skeleton set.
%Now, a cluster $\mathcal{C}_{i}$ contains at least $\frac{\alpha h}{2}$ points that were taken from the corresponding core $\mathcal{R}_{i}$. 
%In this pase we can easily lower-bound the rate in which core points become skeleton points of the corresponding lcuster and
%that turns out to suffice to complete the proof. 
%Below details.

For a set of points $\mathcal{D}$ we say that a point $v$ \textit{dominates $\mathcal{D}$} if the following is true: At least one of the $h$ random numbers assigned to $v$ by the algorithm is 
smaller than all corresponding random numbers assigned to data points from $\mathcal{D} \backslash \{v\}$.
Denote the set of all data points that are outliers as $\mathcal{O}$.
Let us fix a core $\mathcal{R}_{i}$ and some constant $W$.
Denote by $t_{0},t_{1},...$ time stamps at which first $W$, $W+1$,... points from $\mathcal{R}_{i} \cup \mathcal{O}$ are collected.
Let us first find a lower bound on the probability of the following event $\mathcal{F}$:
for every $t_{j}$ in every ball of the $(s,\frac{\Delta}{4})$-covering of $\mathcal{R}_{i}$, 
there are at least $\alpha h$ points from $\mathcal{R}_{i}$ that are dominating the set of all the points collected up
to time $t_{j}$. 
Fix some ball $\mathcal{B}$ of the covering of $\mathcal{R}_{i}$.
By the definition of $\Gamma_{i}$, we know that on average $(W+j)\Gamma_{i}$ 
points from $\mathcal{R}_{i} \cap \mathcal{B}$ arrived up to time $t_{j}$.
By Azuma's inequality, we know that this number is tightly concentrated around its average, i.e.,
for every $\epsilon_{1} > 0$ the probability that for a fixed $t_{j}$ and fixed ball $\mathcal{B}$
this number is less than $(W+j)(\Gamma_{i}-\epsilon_{1})$ is at most $e^{-2W\epsilon_{1}^{2}}$ .
In fact, by using a more general version of the Azuma's inequality, we can get rid of a fixed $t_{j}$
and have the same upper bound for all $t_{j}$s simultanously.
Consider the following event $\mathcal{G}$: for every ball $\mathcal{B}$ up to time $t_{j}$ for every $j=0,1,...$ at least $(W+j)(\Gamma_{i}-\epsilon_{1})$ data points from that ball have arrived. 
Thus, using union bound over all
the balls of the covering, we conclude that $\mathcal{G}$ happens with probability at least $1-se^{-2W\epsilon_{1}^{2}}$. Now we analyze $\mathcal{F}$ conditioned on $\mathcal{G}$.
For any fixed $t_{j}$, the average number of dominating points in the fixed ball $\mathcal{B}$ of the covering of $\mathcal{R}_{i}$ is at least $h(\Gamma_{i}-\epsilon_{1})$. Besides, as previously, one can easily note that the actual number is tightly concentrated around the average. Taking $\delta = 
\Gamma_{i} - \epsilon_{1} - \alpha$ and using Azuma's inequality once again, we derive an upper bound
of the form $e^{-2h\delta^{2}}$ on the probability that a fixed ball $\mathcal{B}$ of the covering
at some fixed time $t_{j}$ contains fewer dominating points than we assumed above.
Using this and taking the union bound over all $t_{j}$ and all $s$ balls of the covering we get:
$\mathbb{P}(\mathcal{F}^{c}) \leq se^{-2W\epsilon_{1}^{2}} + (1-se^{-2W\epsilon_{1}^{2}})sne^{-2h\delta^{2}}$, where $\mathcal{X}^{c}$ stands for the complement
of an event $\mathcal{X}$. 
Let $\mathcal{H}$ denote an event that among first $W$ points from $\mathcal{R}_{i} \cup \mathcal{O}$
there will be at most $Wx$ outliers, where: $x = \gamma_{i}(1+\epsilon_{2})$ and $\epsilon_{2}>0$
is some fixed constant. From Chernoff's inequality we get: $\mathbb{P}(\mathcal{H}^{c}) \leq
e^{-\frac{\epsilon_{2}^{2}}{2+\epsilon_{2}}W\gamma_{i}}$.
Now assume that $\mathcal{H}$ happens and that $W$ points from  $\mathcal{R}_{i} \cup \mathcal{O}$ have already arrived.
Denote by $\mathcal{I}$ the following event: for all balls $\mathcal{B}$ of the covering of 
$\mathcal{R}_{i}$ at any time $t'$ no $\alpha h$ skeleton points in any ball are outliers.
Note that if $\mathcal{I}$ holds then after $W$ points have been seen, at least
$\alpha h-Wx$ new points from $\mathcal{B}$ that are outliers must be seen up to time $t'$. 
Fix again a ball $\mathcal{B}$ of the covering of $\mathcal{R}_{i}$.
Let us take next $t$ points coming from $\mathcal{R}_{i} \cup \mathcal{O}$ for $t=\alpha h-Wx, \alpha h-Wx+1,...$
(we will take $W$ in such a way that $\alpha h-Wx>0$). 
For a fixed $t$ the probability that out of those $t$ points there are more than $(1+\epsilon_{2})t\gamma_{i}$ outliers is, by Chernoff's bound, at most 
$e^{-\frac{\epsilon_{2}^{2}}{2+\epsilon_{2}}t\gamma_{i}}$.
Denote by $\mathcal{J}$ the following event: for all $t \in \{\alpha h-Wx, \alpha h-Wx+1,...\}$ there are at most $(1+\epsilon_{2})t\gamma_{i}$ outliers out of those $t$ points. By the union bound we have: 
$\mathbb{P}(\mathcal{J}^{c}) \leq ne^{-\frac{\epsilon_{2}^{2}}{2+\epsilon_{2}}t\gamma_{i}}$.
Assume that $\mathcal{J}$ holds and fix $t$. The probability that after $t$ new points have been seen,
one of the outliers will become a skeleton point in a fixed ball $\mathcal{B}$ based on the fixed $m^{th}$
random number that was assigned to it by the algorithm is at most 
$(1+\epsilon_{2})\gamma_{i}$. The probability that this will be the case for at least
$\alpha h-Wx$ outliers is at most $((1+\epsilon_{2})\gamma_{i})^{\alpha h-Wx}$. Thus, if we take the union bound
we can conclude that $\mathbb{P}(\mathcal{I}^{c}) \leq n(e^{-\frac{\epsilon_{2}^{2}}{2+\epsilon_{2}}t\gamma_{i}}+s((1+\epsilon_{2})\gamma_{i})^{\alpha h-Wx})$.
Notice that $\mathcal{E}_{i} \subseteq \mathcal{F}^{c} \cup \mathcal{H}^{c} \cup \mathcal{I}^{c}$.
Thus we obtain $\mathbb{P}(\mathcal{E}_{i}) \leq \mathbb{P}(\mathcal{F}^{c})+\mathbb{P}(\mathcal{H}^{c}) + \mathbb{P}(\mathcal{I}^{c})$.
We conclude that $\mathbb{P}(\mathcal{E}_{i}) \leq se^{-2W\epsilon_{1}^{2}} + (1-se^{-2W\epsilon_{1}^{2}})sne^{-2h\delta^{2}}+e^{-\frac{\epsilon_{2}^{2}}{2+\epsilon_{2}}W\gamma_{i}}+n(e^{-\frac{\epsilon_{2}^{2}}{2+\epsilon_{2}}W\gamma_{i}}+s((1+\epsilon_{2})\gamma_{i})^{\alpha h-Wx})$.
Thus, taking the union bound over all the cores we obtain:  $\mathbb{P}(\mathcal{E}) \leq kse^{-2W\epsilon_{1}^{2}} +
ksne^{-2h\delta^{2}}+k(n+1)e^{-\frac{\epsilon_{2}^{2}}{2+\epsilon_{2}}W\gamma_{i}}
+kns((1+\epsilon_{2})\gamma_{i})^{\alpha h-Wx}$.
If we fix: $\epsilon_{1}=\alpha=\frac{\Gamma_{i}}{4}$
and $W=\frac{h\Gamma_{i}}{16 \gamma_{i}}$, then by bounding each term of the RHS expression in the last inequality with $\frac{\epsilon}{k}$, we obtain the lower bound on $h$ as in the statement of the theorem.
Since we have already noticed that it suffices to find an appropriate upper bound on $\mathbb{P}(\mathcal{E})$, this concludes the proof.
\end{proof}

\subsection{Proof of Theorem \ref{rightmergetheorem}}

We will use notation from the proof of Theorem \ref{wrongmergetheorem}.
\begin{proof}
Fix some $\mathcal{R}_{i}$.
Note than we have already proved that an event $\mathcal{F}$ does not hold with probability at most
$se^{-2W\epsilon_{1}^{2}} + (1-se^{-2W\epsilon_{1}^{2}})sne^{-2h\delta^{2}}$. 
Also note that from the definition of $\mathcal{F}$ we know that if $\mathcal{F}$ holds then the number of clusters computed 
by the algorithm and containing at least one point from $\mathcal{R}_{i}$ will not increase after the time when first 
$W$ data points from the corresponding cluster $\mathcal{C}_{i}$ have been seen. 
Let us assume that $\mathcal{F}$ holds. Notice that by the time every ball of the covering gets at least one new data point, 
there will be just one cluster computed by the algorithm with points from the core $\mathcal{R}_{i}$.
When this is the case no further errors regarding new points from the core $\mathcal{R}_{i}$ will be made.
From Azuma's inequality and the union bound we immediately get: the number of extra data points coming from $\mathcal{C}_{i}$ that need
to be seen to populate every ball of the covering with at least one of them is more than $T$ with probability at most
$se^{-2T\Gamma_{i}^{2}}$. 
We conclude that with probability at most $se^{-2W\epsilon_{1}^{2}}+sne^{-2h\delta^{2}}+se^{-2T\Gamma_{i}^{2}}$
after $T+W$ points of $\mathcal{C}_{i}$ have been already seen, the algorithm will still make mistakes on the new data points from $\mathcal{R}_{i}$.
If we now upper-bound every ingredient of the above sum by $\frac{\epsilon}{3}$ and solve for $W$ and $T$, then we get:
$W \geq \frac{8}{\Gamma^{2}}\log(\frac{3s}{\epsilon})$, $T \geq \frac{1}{2\Gamma^{2}}\log(\frac{3s}{\epsilon})$.
Taking the number of points $h$ as in the previous theorem concludes the proof.
\end{proof}

\end{document}